\documentclass[acmlarge,screen,nonacm]{acmart}
\AtBeginDocument{%
  \providecommand\BibTeX{{%
    \normalfont B\kern-0.5em{\scshape i\kern-0.25em b}\kern-0.8em\TeX}}}

\def\inst#1{\unskip$^{#1}$}

\usepackage{lineno}
\usepackage{enumerate}
\usepackage{natbib}
\makeatletter
\let\@authorsaddresses\@empty
\makeatother
\begin{document}
\pagenumbering{arabic}
\title{Taking the Next Step with Generative Artificial Intelligence:\protect\\ The Transformative Role of Multimodal Large Language Models in Science Education}


\author{Arne Bewersdorff\inst{1}}
\authornote{Corresponding author. Email: arne.bewersdorff@tum.de}

\author{Christian Hartmann\inst{1}}

\author{Marie Hornberger\inst{1}}

\author{Kathrin Seßler\inst{1}}

\author{Maria Bannert\inst{1}}

\author{Enkelejda Kasneci\inst{1}}

\author{Gjergji Kasneci\inst{1}}

\author{Xiaoming Zhai\inst{2}}

\author{Claudia Nerdel\inst{1}}

\affiliation{%
  \institution{\newline$^1$TUM School of Social Sciences and Technology, Technical University of Munich}
  \city{Munich}
  \state{BY}
  \country{Germany}
}

\affiliation{%
  \institution{\newline$^2$AI4STEM Education Center \& Department of Mathematics, Science, and Social Studies Education, University of Georgia}
  \city{Athens}
  \state{GA}
  \country{USA}
}

\renewcommand{\shortauthors}{Bewersdorff et al., 2024}

\begin{abstract}
    \textit{The integration of Artificial Intelligence (AI), particularly Large Language Model (LLM)-based systems, in education has shown promise in enhancing teaching and learning experiences. However, the advent of Multimodal Large Language Models (MLLMs) like GPT-4 with vision (GPT-4V), capable of processing multimodal data including text, sound, and visual inputs, opens a new era of enriched, personalized, and interactive learning landscapes in education. This paper derives a theoretical framework for integrating MLLMs into Multimodal Learning. This framework serves to explore the transformative role of MLLMs in central aspects of science education by presenting exemplary innovative learning scenarios. Possible applications for MLLMs range from content creation to tailored support for learning, fostering engagement in scientific practices, and providing assessments and feedback. These applications are not limited to text-based and uni-modal formats but can be multimodal, thus increasing personalization, accessibility, and potential learning effectiveness. Despite the many opportunities, challenges such as data protection and ethical considerations become salient, calling for robust frameworks to ensure responsible integration. This paper underscores the necessity for a balanced approach in implementing MLLMs, where the technology complements rather than supplants the educators' roles, ensuring an effective and ethical use of AI in science education. It calls for further research to explore the nuanced implications of MLLMs for educators and to extend the discourse beyond science education to other disciplines. Through developing a theoretical framework for the integration of MLLMs into Multimodal Learning and exploring the associated potentials, challenges, and future implications, this paper contributes to a preliminary examination of the transformative role of MLLMs in science education and beyond.}

\end{abstract}






\settopmatter{printfolios=true}
\maketitle

\section{Introduction}
Science education encompasses a broad spectrum of activities, from acquiring scientific knowledge and engaging in scientific practices to effective communication about scientific findings and ideas (US: \citet{Council2012}). These activities are essential to prepare students for the complex, multifaceted challenges they will encounter in their future lives \cite{OECD2018} and, therefore, contribute to the 21st-century competencies (e.g., \citet{kay2016framework}; for science education: \citet{McComas2014}). 
The essence of science learning is inherently multimodal, which requires students to engage in science and engineering practices with different modalities: reading and writing scientific explanations or arguments, interpreting diagrams, drawing models, flowcharts, diagrams, analyzing and visualizing data, and designing solutions. These activities facilitate the understanding of scientific knowledge and improve domain-specific competencies \cite{Lemke1998, vanLeewen1990,doran2021teaching}. Additionally, engaging in these activities requires students to seamlessly shift between different modes. Representations such as images, graphics, tables, technical symbols, formulas, equations, process workflows, graphs, videos,  and audio data are essential learning modes in the learning process. This multimodal nature of science learning demands science educators to facilitate learning with multimodal materials and provide students with opportunities to engage in multimodal activities. 

Aligned with the multimodal nature of science itself, research has documented evidence regarding the effectiveness of multimodal learning for fostering learning outcomes in science education. 
"Multimodal learning" (as defined and used herein) has a cognitive foundation based on the multimedia learning theory (CTML) posited by Mayer (e.g., \citet{Mayer1997,Mayer2021}).  Combining multimodal representations like text and images can enhance knowledge acquisition, improving hence the knowledge pieces into a coherent multi-faceted mental model. In alignment with findings from  \cite{Mayer1997},  \citet{Schnotz2003} provided strong evidence that multimodal approaches are effective in promoting (science) learning.  This is because, human beings learn via different information channels and each channel can bear certain amount of working load. Learners struggle when information is intensified and provided in a single modality. Multimodal learning offers opportunities for learners to reduce the cognitive load for each channel. Despite this potential, researchers found that students seldom have the opportunity to engage in high-quality multimodal learning due to the shortage of materials that teachers could generate and provide.


With the latest advances in Artificial Intelligence (AI), specifically in generative AI, Multimodal Large Language Models (MLLMs) have shown great potential to address this gap. MLLMs are a subset of AI models focusing on processing and generating human-like text, images, videos, audio, etc., enabling thus a wide range of applications from content creation to problem-solving \cite{Bommasani2021}. They can also transform unstructured data into structured content and vice versa~\cite{borisov2022language}. 
MLLMs such as GPT-4 or Gemini enable educators to generate learning resources (e.g., lesson plans (Lee et al., 2024)), understand scoring rubrics, and assign scores to students' written responses or drawn models, meeting the needs of multimodality in science education \cite{lee2023nerif}.


In this article, we articulate the potentials and challenges of MLLMs in reshaping the science learning processes and environments. 
Building upon Mayer's CTML, we first develop an AI-enhanced multimodal learning framework and then present a series of exemplary scenarios where MLLMs are used to increase the modalities of learning, including instructional strategies and design, student engagement, assessment, and feedback. We also demonstrate the capacity of MLLMs to provide adaptive multimodal learning support to students with diverse needs.  
At last, we discuss the challenges related to implementing and applying MLLMs in the classroom, such as data protection and ethical concerns, which have become increasingly pressing with MLLMs  


\section {Review of the literature}

\subsection{Core Elements of Science Education} 
Science education is crucial in preparing informed citizens capable of navigating an increasingly complex world \cite{OECD2018}. This education involves imparting a comprehensive understanding of core scientific concepts and crosscutting ideas essential for building a solid foundation in science \cite{Council2012}. Alongside this robust content knowledge, the development of scientific thinking and practical skills is critical. Engaging students in scientific inquiry and hands-on investigations not only cultivates their critical thinking but also equips them with practical abilities essential for scientific practices \cite{Bybee1997,Flick1993}. Moreover, such an approach may also foster their creativity, enabling them to devise effective solutions for ill-structured problems \cite{hathcock2015scaffolding}. Central to science education is also the emphasis on effective communication skills. Students should be able to articulate and communicate complex scientific ideas with clarity and precision \cite{Council2012}.

Central tasks for educators are creating engaging content and empowering students’ learning. In science education, the latter involves not only fostering the acquisition of content knowledge but also encouraging engagement in scientific practices and the communication of scientific findings, insights, and ideas. Assessment and providing effective feedback are crucial tasks for educators. These three central aspects of science education will guide the presentation of the exemplary potentials of MLLMs in Chapter 3.

\subsection{Large Generative AI Models}

\textbf{Large Language Models.} LLMs are a subset of AI models focusing on processing and generating human-like text, enabling thus a wide range of applications from content creation to problem-solving \cite{Brown2020}. Those deep learning architectures are characterized by their capacity to process and interpret context, generate coherent responses, and provide textual content in real time.

Incorporating human feedback through the training of reward models and applying reinforcement learning in the fine-tuning step \cite{Ouyang2022, lee2023rlaif} has significantly enhanced the generation of human-like language that is aligned with social norms and goals and has led to the success of today's popular language models by OpenAI, ChatGPT and GPT-4 \cite{OpenAI2023}. The PaLM models \cite{Chowdhery2022,Anil2023} from Google are setting a focus on a smaller and denser architecture, achieving similar performance. Also, in the open-source landscape, there are strong competitors like Meta’s LLaMA model \cite{Touvron2023,Touvron2023a}, and the Falcon-40B architecture \cite{Penedo2023}. These models have sparked further development, with researchers fine-tuning them and creating open-source variants like Vicuna \cite{Zheng2023} and Alpaca \cite{alpaca2023}.

The impact of these LLMs has been profound and led to novel advances across various industries. They have enabled innovative approaches in areas such as healthcare \cite{Chintagunta2021,Enarvi2020}, finance \cite{Dowling2023}, journalism \cite{Pavlik2023}, creative writing \cite{Yuan2022}, and even in scientific discovery~\cite{romera2023mathematical}, demonstrating the strength and versatility of LLMs. The advances in the generation of human-like texts have opened up new opportunities and challenges in different domains \cite{Kaddour2023}. This includes a meaningful and constructive application in educational contexts, considering remaining issues like correct mathematical reasoning \cite{Imani2023}, or possible hallucinations in the output \cite{Ji2023}. Research continues to address these challenges while also improving their explainability to increase trust and reliability in the model outputs \cite{Wu2023b}. While LLMs have the potential to significantly impact text-based learning (e.g., \citet{bewersdorff2023assessing, Sessler2023, Kuechemann2023}), the advent of MLLMs promises to extend these benefits to include visual, auditory, and other sensory data modalities.
$ $\\   

\noindent\textbf{Multimodal Large Language Models.} 
Building upon the foundation of LLMs, MLLMs are a further advancement considering their ability to process and generate content across various data types, including text, images, audio, and video \cite{Bommasani2021}.  However, these models are built to understand, interpret, and respond to multiple modalities, enriching context information beyond merely text-based capabilities.
Much initial research focuses on the combination of visual and textual understanding as a multimodal input, yet they predominantly generate output in text form only \cite{Alayrac2022,Zhu2023,Huang2023a,Ye2023}. To accomplish this joint understanding of textual and visual input, developers often integrate existing image encoders with pre-trained LLMs. They enable this combination by incorporating additional modules that project the different data forms into a shared embedding space \cite{Li2023,Liu2023,Zhang2023b}. 

However, the capabilities of MLLMs are not limited to images and text; they can also process combinations of text and video data \cite{Li2023,Maaz2023}. For example, this can be achieved by dissecting video into audio and a series of images, then encoding and embedding each element separately to capture the context of the video format \cite{Zhang2023b}. Taking a more comprehensive approach, multimodal inputs can also move beyond mere dual-modality and include additional sensor data \cite{Driess2023} or speech \cite{Chen2023}. An example of this advancement is PandasGPT \cite{Su2023}, which integrates six different modalities as input for LLMs.

Despite the proficiency of LLMs in processing multimodal context, they are currently mostly designed to output information solely in textual format. By combining visual foundation models \cite{Wu2023}, audio foundation models \cite{Huang2023a}, or both \cite{Shen2023}, LLMs can act as a sophisticated 'controller' between data types. This allows them to intelligently invoke specific AI models as needed, marking a preliminary step toward generating multimodal outputs.

Enhancements in multimodal capabilities, such as more advanced structures for audio processing \cite{Zhang2023}, have the potential to significantly boost performance. The concept of a singular model capable of handling text, images, audio, and video for both input and output is still relatively nascent. 
Recently, Wu et al. published NExT-GPT as a general-purpose any-to-any open-source MLLM \cite{Wu2023a} and OpenAI released GPT-4 Vision \cite{OpenAI2023} and GPT-4 Turbo\footnote{https://chatopenai.de/gpt-4-turbo/}, marking an important further development in this research. Also, Google announced an new LLM-based AI system they named Gemini, which is, according to Google "built from the ground up for multimodality - reasoning seamlessly across text, images, video, audio, and code" \cite{Deepmind2023}. This kind of reasoning in combination with other types of deep learning and explainability approaches~\cite{wu2020comprehensive,borisov2022deep,rombach2022high} has the potential to advance science~\cite{wong2023discovery} and education. Often, these AI systems are labeled Multimodal Foundation Models \cite{li2023multimodal}. Following \citet{fu2023mme}, we will use the term 'Multimodal Large Language Models' in our paper.

\subsection{Adaptive Multimodal Learning} 

Information processing occurs through various channels, each handling different forms of information, such as spoken or written language or images. The dual nature of information (verbal vs. non-verbal) and the associated (specific) processing of different information types is called multimodality. The empirically well-supported idea that combining multiple modalities can enhance knowledge acquisition is documented in Mayer's theory of multimedia learning \cite{Mayer1997, Mayer2021} as well as in Schnotz \& Author's theory of text-picture integration \cite{Schnotz2003}. Based on cognitive models of information processing, such as Paivio's Dual Coding Theory \cite{Paivio1991}, the so-called multimedia effect is the well-supported hypothesis that active processing of both verbal (textual) and non-verbal (visual) information enhances integration into a coherent mental model (for a recent meta-analysis of the multimedia effect, see \citet{Hu2021}). Knowledge acquisition becomes more complete and less ambiguous and, thus, more effective when learners construct mental models from different modalities. Different representations can complement each other or even lead to a better overall understanding. For example, an image of a human skeleton or the depiction of an enzymatic reaction could be more effective and easier to understand than a written description, particularly when considering the time required to process the information. In summary, combining textual and pictorial information can effectively support learning. This can be further specified by the Cognitive Theory of Multimedia Learning (CTML).

According to CTML, multimodal representation enables the selection, organization, and integration of learning content into a coherent mental model \cite{Mayer2021}. Learners first select relevant information from the presented material to focus their attention on. The selected information is then organized in the learner's mind into coherent structures, and finally, learners integrate the organized information with their existing knowledge to make sense of it. For instance, according to the signaling principle (cf. \citet{Gog2014}), learners often have difficulty selecting important information in learning materials, so they need to be supported in their search to avoid unnecessary cognitive load, especially in text-image relations (for a meta-analysis of different studies on the signaling principle, see \citet{Richter2016}). Previous research has shown that signaling, i.e., separate selection aids, can hinder learning, especially for learners with at least sufficient prior knowledge (e.g., \citet{Richter2019}). 

Second, learners have to organize their learning materials, i.e., text and images, according to their needs to keep their cognitive demands as low as necessary for efficient learning processes to take place \cite{Mayer2003}. Third, according to \citet{Mayer2021}, the integration of multimodal representations requires generative activities on the part of the learner. It is, therefore, essential that multimodal representations are actively transformed into coherent mental models by the student. 

If we summarise the processes of selection, organization, and integration and discuss them in the context of MLLM, the potential for adaptivity becomes clear. In multimedia presentations, visual and textual information is usually statically embedded in a media presentation (e.g., a textbook or a graph). However, neither text nor visual information can be easily adapted to the learner's needs in real-time. This changes with MLLMs as they allow the learner to convert textual information into images and vice versa with simple prompts. For example, it is possible to convert text from MLLMs into mind maps, graphical illustrations, or to extract terms from a complex representation of scientific processes, making it much easier for the learner to select, organise and integrate information into complete mental models. In other words, MLLM offer a promising way of "representational scaffolding"  \citep{fischer2022representational} to extract terms from complex scientific representations, helping learners to select and organize, i.e., the adaptation of representational modalities to learners' needs.

As will be discussed below in relation to specific science education contexts, MLLMs offer a potential added value to this three-step process by enabling learners and educators to adapt multimodal representations themselves. As the CTML refers to the processing of multimodal information in the context of teaching-learning processes (cf. \citep{Mayer2021}), it provides a suitable conceptual framework for the following discussion of the potential of MLLMs in (science) education.

\section{Framework of  Integrating MLLM into Multimodal Learning} 
\begin{figure}
    \centering
    \includegraphics[scale=0.6]{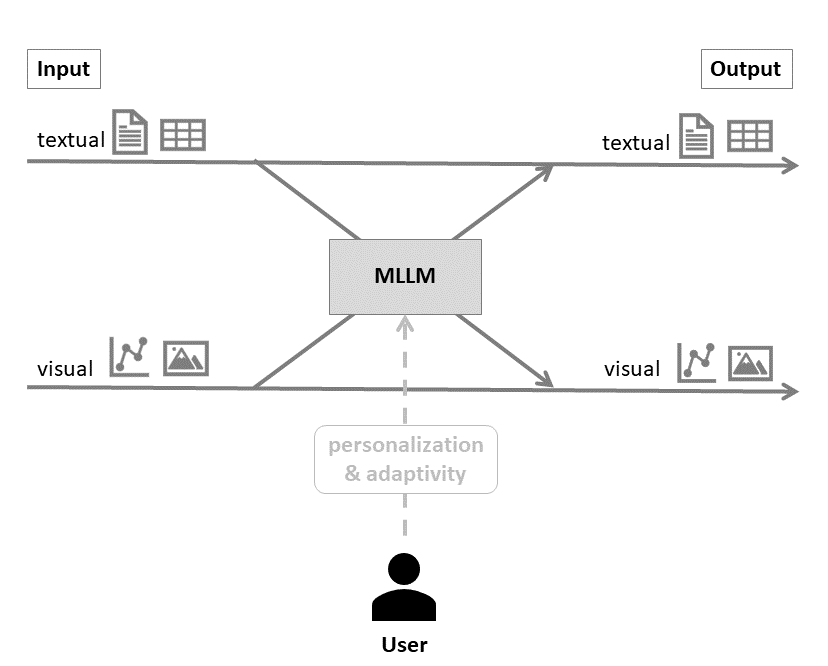}
    \caption{Framework of  Integrating MLLM into  Multimodal Learning}
    \label{fig:framework_MM_AI}
\end{figure}

Rooted in the Theory of Multimedia Learning (cf. 2.3) and considering the capabilities of MLLMs (cf. 2.2), we propose a framework that integrates MLLMs into multimodal learning. Positioned between the verbal (textual) and non-verbal (visual) channels (refer to Section 2.3 and Dual Coding Theory \cite{Paivio1991}), the MLLM operates as an essential processing unit that bridges these channels.
In our framework, the MLLM processes visual or textual input based on the user's requirements for adaptiveness and personalization, producing textual and/or visual outputs. The MLLM has two key functionalities (cf. \autoref{fig:framework_MM_AI}): 
\begin{enumerate}[A.]
    \item Transforming content from text to image or vice versa: MLLMs facilitate the shift between textual and visual representations. For instance, in science education, converting tabular data into visual diagrams can provide further insights. In \autoref{fig:framework_MM_AI}, this example would be represented by the path from textual input through the MLLM to the visual output.
    \item Shifting from uni- to multimodality by adding a modality: As each channel has a limited capacity for processing information at any given time, adding a modality —supplementing text with visuals or vice versa—aims to reduce this cognitive load. Knowledge acquisition becomes more complete, less ambiguous, and thus more effective when learners construct mental models using various modalities (see 2.3).  In science education, one example could be to augment the structure of an insect's eye given in an illustration by textual information. In \autoref{fig:framework_MM_AI}, this example would be represented by both the path from the visual input through the MLLM to the textual output as well as the path from the visual input right to the visual output.
\end{enumerate}

Depending on the setting, the user of the MLLM can be either the educator or the learner itself. With the MLLM, the user can generate adaptive and personalized multimodal representations. When the educator operates the MLLM, they must provide information such as competencies, student needs, and individual difficulties to the MLLM. If the student operates the MLLM on their own, information to ensure adaptivity and personalization must be provided beforehand by the educator, the student themselves, or the MLLM may indirectly retrieve this information through interaction. Adaptability ensures that the transformation and augmentation of modalities are tailored to meet the learner's needs and competencies.
It is crucial for the MLLMs to inherently apply effective multimedia design principles \cite{mayer2024past}, e.g., arising from the signaling principle \cite{Gog2014}. This proposed framework represents an initial draft, laying the groundwork for further development and refinement in integrating multimodal learning with advanced capabilities of MLLMs.

\section{Applications of Multimodal LLMs  for Science Education
}
The following section is structured by central aspects of science education (see 2.1). It aims to present exemplary potentials of MLLMs in science teaching and learning, focusing specifically on adaptive, multimodal learning. The given examples will be discusses in the light of our derived framework (see 3.). The examples discussed in the remainder of this section are succinctly represented in ~\autoref{tab:science_education}.

\begin{table}[h!]
\centering
{\small
\begin{tabular}{|p{0.2\linewidth}|p{0.7\linewidth}|}
\hline
\textbf{Subsection} & \textbf{Exemplary Aspects}\\
\hline
3.1 MLLMs for Content Creation & 
- Tailoring multimodal learning materials to diverse student needs

- Organizing content effectively to reduce cognitive load

- Promoting active engagement through generative activities

- Flexible adaptation of visual representations for easier recognition (signaling)

- MLLM-based code and content generation for virtual reality environments

- Comprehensive integration of MLLMs into virtual reality settings via APIs\\
\hline
3.2 MLLMs for Supporting and Empowering Learning & 
\textbf{3.2.1 Fostering Scientific Content Knowledge}

- Transforming/supplementing text with visuals

\textbf{3.2.2 Fostering the Uses of Scientific Language}

- Assisting in understanding and using scientific language

- Simplifying technical language and explaining technical terms

\textbf{3.2.3 Supporting Engagement in Scientific Practices}

- Assisting in formulating research questions and hypotheses

- Visualizing and interpreting raw data

- Providing contextual explanations alongside visual representations.

\textbf{3.2.4 Supporting Scientific Communication \& Presentation}

- Converting data structures for effective communication

- Generating image-based storyboards from analogies of scientific phenomena\\
\hline
3.3 MLLMs for Assessment and Feedback & 
\textbf{3.3.1 Visual Assessment}

- Personalized assessment of text and visual content in students' reports

- Enhancing quality and objectivity of assessments

\textbf{3.3.2 Multimodal Feedback}

- Providing elaborated feedback with visual aids

- Instant feedback on various modalities like texts and drawings\\
\hline
\end{tabular}
}
\caption{Exemplary aspects of the proposed framework for science education using MLLMs.}
\label{tab:science_education}
\end{table}

\subsection{MLLMs for Content Creation}

The development of learning materials is a central task for educators. These materials should be designed to meet the learners' needs while keeping the learning goal(s) in mind. Multiple representations can foster student motivation, interest, and conceptual understanding \cite{Treagust2008}. For example, a diagram can provide learners with a way of visualizing the concept and hence developing a mental model  \cite{Gabel1998}. The creation of effective learning materials, therefore, often entails a shift in the modality - especially when the educator has to cater materials to diverse learning groups. MLLMs can help educators create tailored, multimodal learning materials to meet the diverse needs of the students.

While more skilled students might work on plain tabular data, others might be served better with a diagram or even (additional) textual or auditive explanations. The educator could provide accompanying visual information to otherwise text-only knowledge representations. For example, this could be done to map certain habitats, give images of listed laboratory equipment, provide images of certain mammals or plants as well as provide structural visualisations of atoms and molecules. Using MLLMs, images can be flexibly adapted to make it easier for learners to directly recognize essential aspects of the visual representation (cf. signaling). 

In addition, MLLMs can help organize learning content effectively, thereby reducing cognitive load, as \citet{Mayer2003} suggested. The organization is crucial for creating materials that are accessible and understandable to a wide range of learners. For example teaching complex organic chemistry reactions, like the mechanism of S\textsubscript{N}1 and S\textsubscript{N}2 reactions, can be challenging due to their abstract nature. MLLMs could generate detailed, step-by-step visualizations of these mechanisms. Due to the adaptive nature of MLLMs, students can manipulate variables and see the effects on the reaction process. 

Fostering engagement with such learning materials via MLLMs through generative activities, integrating multimodal representations \cite{Mayer2021}, could even go further, shifting traditional content creation partially from the educator to the student: The student adaptively creates (organizes) his own learning materials based on an initial impulse, such as a textual description of the mechanism of S\textsubscript{N}1 and S\textsubscript{N}2 reactions - this can range from figures and diagrams to animations, or even synthetic datasets to investigate the chemical reaction.

Viewing the examples through the lens of the framework introduced in section 3, all the provided examples perform a shift from uni- to multimodality by adding a modality. One existing modality is extended with a second modality, potentially reducing cognitive load. In the examples given, the user of the MLLM is either the educator or the student themselves. Both can create adaptive learning materials.

Moreover, MLLM-based learning content can be integrated into innovative, immersive virtual-reality learning environments. The ability of MLLMs to reason about spatial relationships, coupled with their ability to generate relevant code and visual elements, facilitates their use in the design, creation, and extension of virtual environments. In addition, well-defined application programming interfaces (APIs) to MLLMs allow for a comprehensive integration of MLLMs into these virtual settings (see \citet{hartmann2023imagine}).

\subsection{MLLMs for Supporting and Empowering Learning
}
\subsubsection{Fostering Scientific Content Knowledge Construction}
Engagement in real-world science materials is key to an authentic learning experience. Authentic learning can help boost motivation  \cite{Banas2014} and science understanding \cite{nathan2014foundations}. Learning with real-world materials, like research papers or even Wikipedia articles, implies they were not originally designed (primarily) for learning; many original scientific sources are predominantly text-based.

\begin{figure}[htp]
\begin{center}
\includegraphics[scale=0.4]{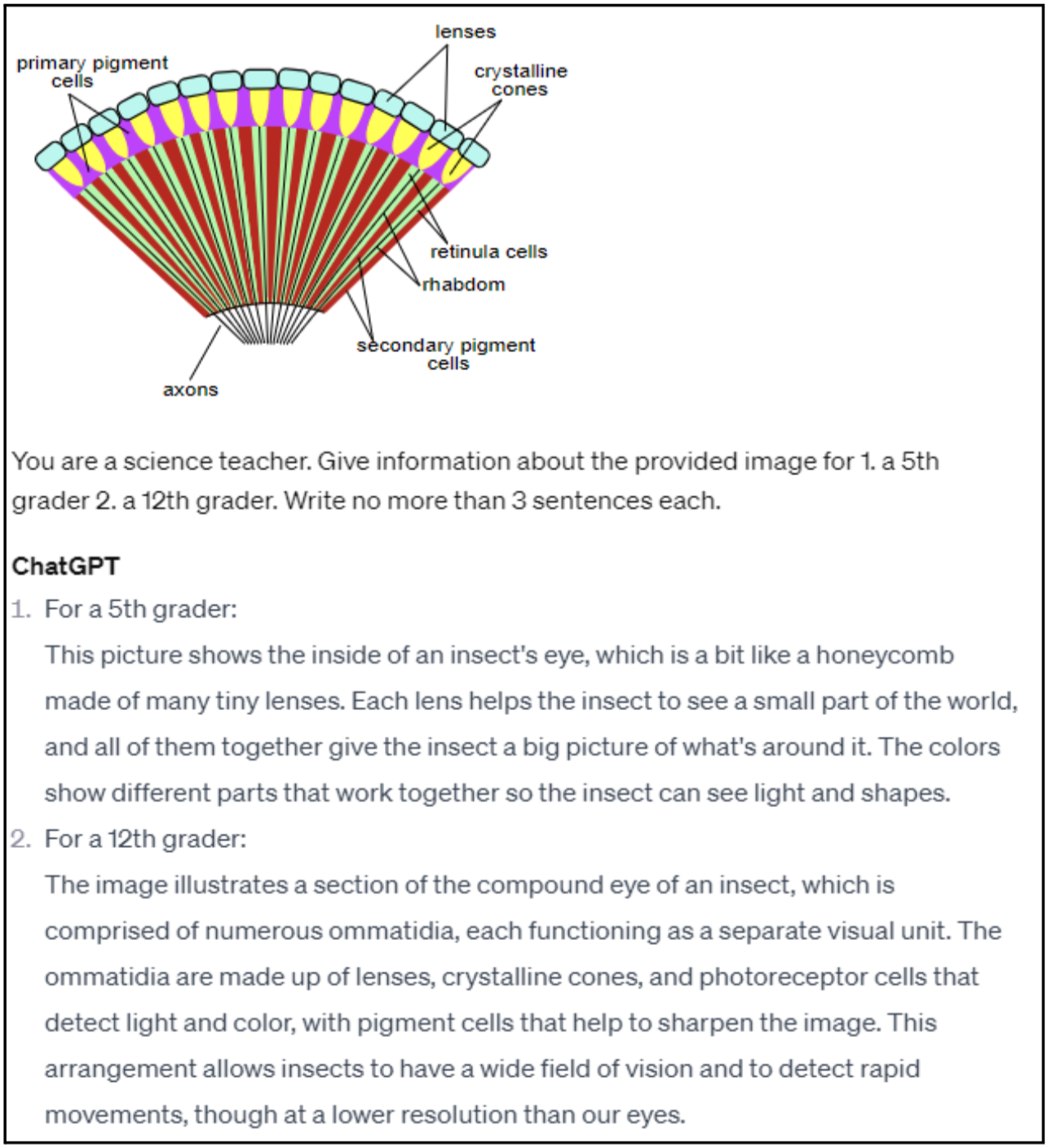}
    \end{center}
    \caption{Example of fostering scientific content knowledge: Graphical representation of an insect's eye uploaded to an MLLM (ChatGPT with GPT4-Vision, \cite{openai2023gpt4v}) asking for explanatory information for a 5th as well as a 12th grade student.}
    \label{fig:intersect_eye}
\end{figure}

MLLMs can enhance these materials by supplementing textual information with visual models (see Chapter 3), thereby aiding students in visualizing complex scientific concepts and increasing the accessibility of original sources. This gives students the opportunity to engage with real-world materials of their choice.
For example, students try to understand the function of an insect's compound eye based on an image they found on Wikipedia. MLLMs could help students to adaptively interact with these materials and complement text-based information (\autoref{fig:intersect_eye}) based on their background and prior knowledge. As exemplarily depicted in \autoref{fig:intersect_eye}, the MLLM is capable of analyzing the image and providing adaptive information tailored to the competency of the learner (i.e., grade 5 and grade 12 student).

It is well known in previous works in education research that learners often face difficulties in understanding and interpreting diagrams \cite{Henderson1999}. Therefore, it is crucial to keep learners' cognitive demands as low as necessary for efficient learning processes to take place \cite{Mayer2003} and further to keep the text and image information understandable for the learner (cf. “pretraining principle”; e.g.,  \citet{Jung2019}). In this context, MLLMs could assist in the reception and understanding of diagrams, making complex data more accessible. This could be achieved by augmenting the diagram with textual explanations on the level of the diagram itself, like the explanation of the labels of the axis or even textualizing the presented data. 

Both examples, when viewed through the framework introduced in Section 3, demonstrate a shift from unimodality to multimodality by incorporating an additional modality. The visual modality, which includes a graphical representation of an insect's eye or a diagram, is augmented with textual information. The user of the MLLM can be in both examples either the learner or the educator.

\subsubsection{Fostering the Uses of Scientific Language}
Learning scientific language is a major aspect of science education  \cite{Wellington2001,Nielsen2012}. But non-scientists, like students, often find scientific language to be severely limiting and difficult \cite{Gardner2012}. From the perspective of multimedia learning, it is essential that the text and image information is understandable to the learner \cite{Jung2019}. MLLMs could be used in this context to help students recall and remember scientific terms in images or diagrams. Using MLLMs, learners can adaptively augment textual information like technical language, or have technical terms explained on the fly, for example, from a scientific podcast.
Previous unimodal learning material, visuals such as images or diagrams, are supplemented by a second modality, textual information like technical language (cf.  Section 3).

\subsubsection{Supporting Engagement in Scientific Practices
}
Engaging in scientific practices, such as modeling or planning and executing experiments, are central components of science curricula in Germany  \cite{KMK2004} and around the world (e.g., US: \citet{Council2012}, UK: \citet{department2013national}, and globally: \citet{OECD2018}). MLLMs have the potential to assist students in various scientific practices, from generating ideas and models to interpreting diagrams and guiding investigations. 

When confronted with a scientific problem, students often struggle to formulate adequate research questions eligible for scientific investigation \cite{Neber2008,Hofstein2005}. MLLMs can inspire students to think about potential scientific questions and help in the process of formulating adequate hypotheses. For example, an educator could provide an image of an eclipse, asking students to formulate questions they want to investigate. If some students struggle to develop adequate hypotheses on their own, the educator has to provide appropriate scaffolds like providing more images, some complementary textual information, or motion graphics like videos and animations of the eclipse to prevent stagnation in the students’ scientific investigation. MLLMs could help generate appropriate scaffolds in different modalities on-the-fly. Equipped with tailored scaffolds, it still remains up to the educator to choose the appropriate degree of guidance.

Furthermore, visualizing and interpreting raw, unstructured data often challenges many students (e.g., \citet{Kotzebue2014, Setiawan2021}). MLLMs have the capability to render raw data into diagrams. Furthermore, MLLMs could significantly enhance understanding by explaining the key properties of the visualized data. These properties might include information about the x- and y-axes, underlying mathematical functions, or special points of interest within the data. MLLMs can provide tailored support by potentially shifting or adding another modality to the mathematical or data level, facilitating thus a deeper understanding. Moreover, they can assist in interpreting the data or drawing conclusions, thereby aiding in a comprehensive understanding of the data's significance. By providing contextual explanations alongside visual representations, MLLMs can bridge the gap between raw data and meaningful interpretation, aiding not only in comprehension but also in deriving interpretations and conclusions from the data.

\begin{figure}[htp]
\begin{center}
\includegraphics[scale=0.5]{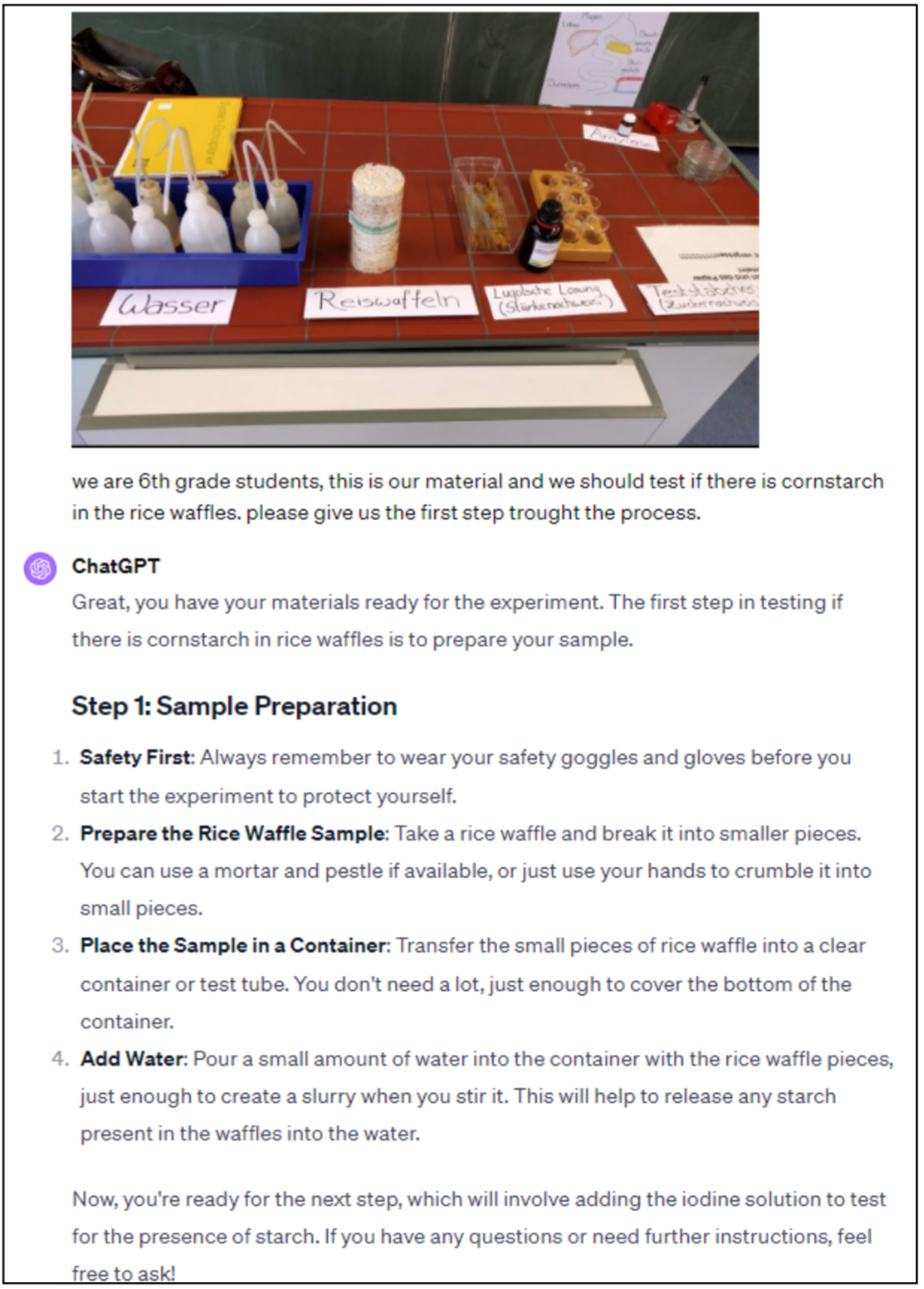}
    \end{center}
    \caption{Example of fostering scientific practices: Students can ask about which steps to perform in an inquiry by uploading the given material (here, ChatGPT with GPT4-V \cite{openai2023gpt4v} was used).}
    \label{fig:gpt_chat}
\end{figure}

Engaging with specialized laboratory equipment is an essential aspect of students' scientific inquiry processes, but instructions often refer to laboratory equipment in technical terms. Additionally, students sometimes also have difficulties with properly handling measuring and laboratory equipment \cite{Kechel2016}. In this context, MLLMs can offer images of unknown instruments, preventing thus misuse and confusion. Beyond the static augmentation of names of laboratory equipment by explanatory images, MLLMs hold the potential to guide students during the use of laboratory equipment. By capturing photos of laboratory equipment with devices like smartphones or tablets and detailing the issues faced during operation, MLLMs can effectively guide students through the process, adapting responsively to their needs (see \autoref{fig:gpt_chat}).

Note-taking is crucial for the success of scientific inquiry, yet it can be a complex task. To be successful, students first need a basic understanding of what and how to write down their notes  \cite{MacNeil2010}, which draws attention and ultimatively cognitive load during the process of scientific inquiry. MLLMs can help by transcribing spoken words into text summaries, which aids in note-taking without drawing any attention and cognitive capacities away from the scientific inquiry. Educators must always carefully consider how to provide scaffolding in their teaching without leading students to, at worst, mindlessly follow procedures and outsource the actual thinking to the MLLM. Reducing cognitive load can be especially beneficial for introductory science courses.

All these examples demonstrate the variety of how MLLMs can be employed in scientific practices, always aligning with basic multi-modal learning theories: multimodal aids can help students grasp complex scientific concepts and formulate relevant questions, which aligns with  \citet{Mayer2021} and  \citet{Schnotz2003} theories on the integration of text and images, facilitating the construction of mental models from different representations. In our framework (Section 3), this process relates to augmenting one modality with a second modality. The user can be either the educator or the student. Based on information provided by the user, the MLLM offers adaptive learning materials in a second modality.

\subsubsection{Supporting Scientific Communication \& Presentation}
\begin{figure}[htp]
\begin{center}
\includegraphics[scale=0.45]{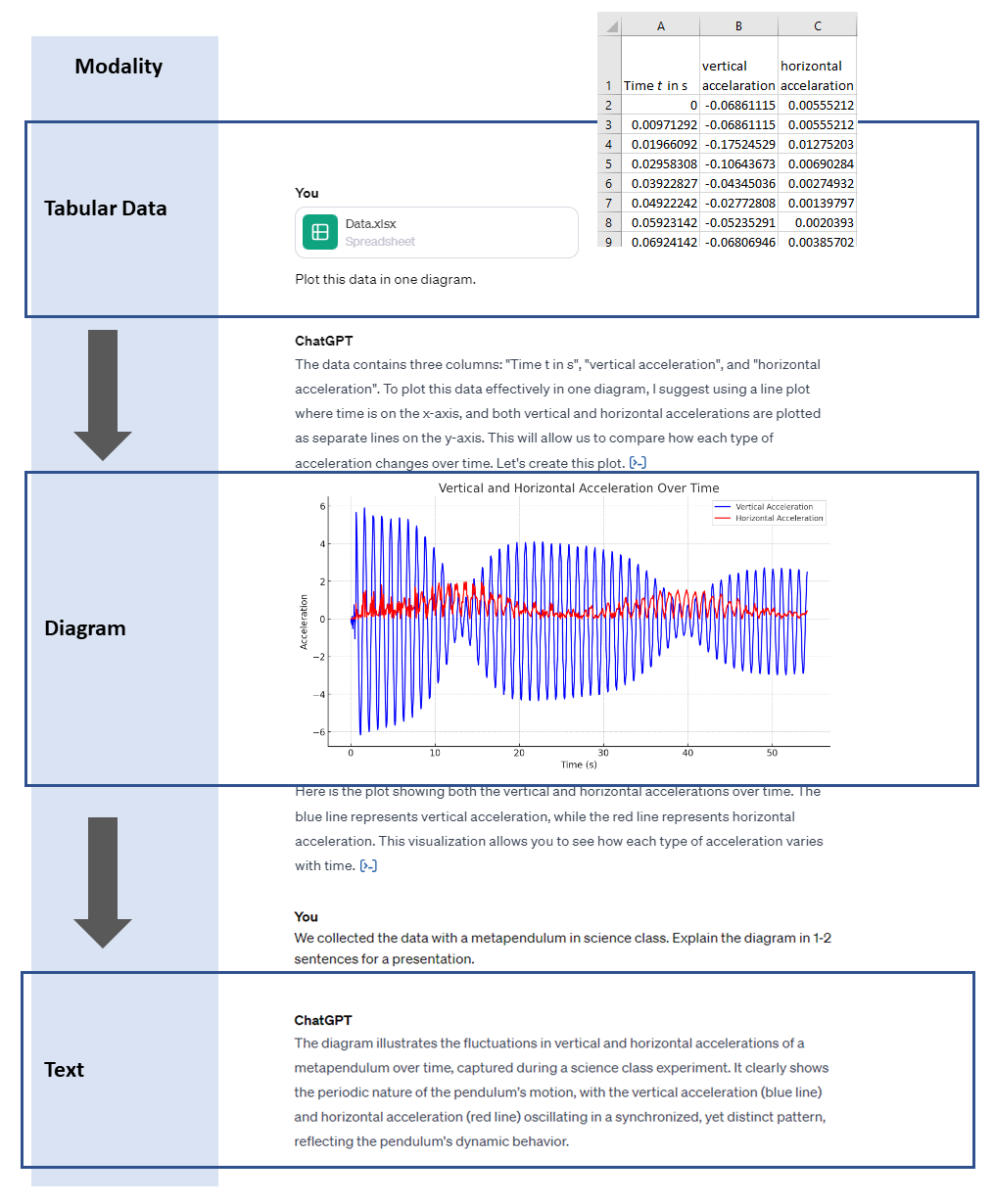}
    \end{center}
    \caption{Example of supporting students' scientific presentations: Students can adaptively use their tabular data to create diagrams and then generate explanatory text based on these diagrams.}
    \label{fig:gpt_pres}
\end{figure}

Communication of scientific results and ideas is a central aspect in science curricula \cite{department2013national,Council2012,KMK2004}. For students to proficiently convey and articulate scientific concepts, they must generate new content. Therefore, it is important that they actively integrate new information with their pre-existing knowledge to make sense of and create their content \cite{Mayer2021}. This integration of multimodal representations requires generative activities like creating diagrams from data by the student. MLLMs can assist in this process by converting data structures, such as tabular data, into diagrams, animations, or graphics, and transforming graphics into text, thereby supporting the process of creating content for scientific communication. One example is provided in \autoref{fig:gpt_pres} where a shift from tabular data to a diagram and, finally, textual data is demonstrated.

Another example is the adaptive generation of complete image-based storyboards building on given analogies of scientific phenomena, as demonstrated by \citet{cao2023elucidating} (example: ibidem). In this process, firstly, abstract scientific concepts are translated into relatable text-based analogies to simplify understanding; secondly, these analogies are further converted into static visual representations to enhance conceptual clarity; and finally, these static visuals are evolved into dynamic visual analogies, offering dynamic representations.

In our framework (Section 3), the creation of diagrams and then generating explanatory text  based on tabular data relates to transforming content from one modality to a second modality as well as augmenting one modality with another. Tabular data (textual) is transformed into a visual representation (diagram). This shift in modalities provides further insights into the data. In the second step, the diagram is augmented by textual information describing the diagram. Based on information provided by the user, the MLLM offers adaptive learning materials in a second modality.

\subsection{MLLMs for Assessment and Feedback}
\subsubsection{Visual Assessment}
Assessment is crucial for successful learning trajectories  \cite{Hattie2008,Hattie2007}. Especially forms of (computer-based) ‘rapid formative assessment’ have shown to be highly effective  \cite{Yeh2010}. 

Visual representations in science education encompass a wide variety of elements, including, but not limited to, mathematical graphs, diagrams, figures illustrating models and processes, as well as students' sketches from experiments. They serve as various channels to convey and assimilate scientific ideas. However, there is a noticeable lack of using the full potential of assessing these visual representations. This primarily stems from the substantial time and effort required on the part of educators to evaluate these often very complex, incomplete, or even contradictory visual constructs. 

There are already many tools that use LLMs to provide assessment in science education, e.g., \citet{bewersdorff2023assessing, Zhai2023, latif2024fine}. All of these mentioned assessment tools provide textual information based on textual input; they are uni-modal. Therefore, the ways of assessment are limited, especially in science education, with their predominant multimodal representations. MLLMs can deliver on any modality, enhancing the possibilities of assessment. 

Students often create written reports on an experiment, including drawings of their experimental setup. MLLMs, with their capability to assess text and visual content like graphs and diagrams, could provide a comprehensive, personalized assessment of their written reports, ultimately providing a deeper understanding of students' cognitive processes and their skill in conveying complex science in both written and visual modes. Besides analyzing students' cognitive processes by using MLLMs for assessment, educators could potentially save time and enhance both the quality and objectivity of assessments. \citet{lee2023nerif} demonstrated an AI system assessing drawn models of scientific phenomena, pioneering new ways of visual assessment. For visual assessment, the visual input, such as a student's drawing, is transformed into a textual representation, such as written feedback or a written assessment (cf. section 3).

\subsubsection{Multimodal Feedback}
Feedback, especially if elaborated and not only providing the correct answer, has been shown to be highly effective for students' learning  \cite{VanderKleij2015}. Building on successful uni- or multimodal assessment, feedback can make student learning more effective. As with assessment, current feedback tools are mostly text-based e.g., \citet{Sessler2023} and, therefore, might not use their full potential to provide effective feedback in science education. With regard to CTML \cite{Mayer2021}, multimodal feedback, combining visual and textual elements, engages ideally dual-channel cognitive processing of learners. This integration not only balances cognitive load but also enhances comprehension and retention, ultimately fostering understanding. 

Students could receive feedback complemented by visual aids for their textual work. For example, visual representations like models of atoms, molecules, and their reactions could be provided to students describing a chemical reaction, helping them clarify misconceptions. Due to the responsiveness of MLLMs, these figures do not have to be stored and manually annotated to the particular misconceptions as in common, non-MLLM-based feedback tools.

The design of feedback should always be considered in relation to subsequent related tasks \cite{henderson2019conditions}. MLLMs possess the capability to provide almost instant feedback, not only on student texts but also on other modalities like their drawings. This immediate response enables students to promptly adjust their work, thereby preventing them from pursuing ineffective approaches that could halt their learning process. For multimodal feedback, traditional textual feedback can be augmented with an additional visual modality like a diagram. The user of the MLLM  can be either the educator or the student, as feedback may be directed to the student for personal reflection or to the educator for further refinement or validation. Regardless of the recipient, feedback can always be personalized and adaptive by the MLLM (cf. section 3).

\section {Challenges and Risks of MLLMs in Science Education} 
While the adaptability of MLLMs in science education is promising, it is essential to approach their potential with caution. The allure of providing students with extensive flexibility in choosing the desired modalities within their learning environment can be strong. However, evidence suggests that minimal guidance in science education may not always yield optimal learning outcomes \cite{Kirschner2006,Sweller2007}. Learners might require guidance in selecting the most appropriate modality to enhance their educational experience effectively. An abundance of options can not only serve as a distraction, but can also elevate the overall cognitive load, potentially hindering the learning process. This is especially true for students with low self-regulation capabilities  \cite{Lan1998,Ley1998}. Consequently, the roles of educators remain pivotal in guiding the learning process \cite{ley2001instructional}. They play a critical role in facilitating the effective use of these advanced tools, ensuring that the technology enhances, rather than impedes, the learning experience. In this context, the nuanced integration of MLLMs calls for a balanced approach, where the technology is leveraged thoughtfully to complement and augment traditional educational practices rather than replace them outright.

Students should engage in scientific practices in problem-based learning and open inquiries \cite{Council2012}. But as minimal guidance can hinder effective learning in science education \cite{Kirschner2006, Sweller2007} MLLMs should be designed to be adjustable by the educator to the individual student’s competency level. In this way, MLLMs would be able to provide degrees of guidance or scaffolding rather than acting as a personalized, adaptive, and multimodal agent that provides immediate answers and solutions. This mediated approach could encourage students to engage in critical thinking and problem-solving, fostering a deeper understanding and promoting self-directed learning without overexerting the student. The MLLMs should be aligned with educational and instructional strategies that prioritize the development of students’ analytical skills in tandem with their knowledge acquisition.

Ethical considerations encompass the moral and ethical implications of using MLLMs, including issues of bias, privacy, consent, and the overall impact on the quality and equity of education. In the few examples we provided, the MLLMs worked quite satisfactorily, potentially improving in the future. However, it is important to acknowledge that MLLMs sometimes still generate biased \cite{venkit2023nationality} incorrect, faulty or fabricated content (often referred to as hallucinations)~\cite{ji2023survey,zhang2023siren,azamfirei2023large,manakul2023selfcheckgpt}, particularly in the context of logical and mathematical reasoning \cite{imani2023mathprompter}. Therefore, it is crucial for researchers to empirically examine the potential bias that may contaminate assessment validity \cite{Zhai2023Ross}.  Research has already shown evidence that AI has the potential to enlarge the score divergence by gender or English language proficiency \cite{latif2023ai}. Educators need to be aware of these limitations, as well as over-blow of AI bias (e.g., pseudo AI bias, see \cite{zhai2022pseudo}). Moreover, if students are encouraged to use MLLMs independently, it is essential that educators communicate these types of potential flaws to them. Fostering a culture of ethical data handling among educational stakeholders, alongside creating awareness about the potential risks and safeguards among the student populace, is crucial to ensure a secure and responsible utilization of MLLMs in educational settings. 

With AI becoming an increasingly integral part of our daily lives, policymakers have begun to implement policies to regulate and safeguard AI development and its use in education \cite{schiff2020s}. E.g., the currently presented so-called European AI Act (\citet{eu2023aiact}) requires generative AI systems, such as MLLMs, to disclose when content is AI-generated, design models to prevent the generation of illegal content, and publish summaries of copyrighted data used for training. The use of such AI systems, and therefore MLLMs, in education will be classified as high-risk applications and will require registration in an EU database. On every level of deploying MLLMs, all stakeholders, such as developers, educators, administrators, and policymakers, as well as the corresponding governance framework, have to prioritize ethical and regulatory considerations. 

Educators and learners have a variety of pre- and misconceptions, fears, and hopes about AI  \cite{Bewersdorff2023}, which might lead to general skepticism toward AI systems in the classroom among stakeholders \cite{Douali2022}, ultimately hindering its effective implementation. For effective implementation, policymakers do not only have to ensure that all learners and educators have access to these potentially powerful tools for learning - they have to provide them with knowledge about AI (AI literacy: \citet{Long2020, Hornberger2023}) and the competencies they need to successfully employ these AI systems. 

Looking forward, there are also risks associated with proprietary MLLMs, such as unknown ethical considerations and implementations. Hence, the decision between open-source and proprietary MLLMs becomes pivotal in the long term. While proprietary models may offer robust data security and professional support, they often come with limitations in terms of accessibility, customization, and cost. This could hinder collaborative learning and innovation, as proprietary MLLMs might not be as adaptable or transparent as their open-source counterparts. Moreover, through restrictions in the ability to tailor proprietary models to specific educational contexts, such models might not always align with the diverse needs of educators, learners, researchers, or other stakeholders. As technology advances, the choice between open-source and proprietary MLLMs may not only be decisive for the trajectory of adoption of generative AI in education but also the realization of an inclusive, digitally advanced educational landscape.

\section{Discussion}
LLMs have emerged as a transformative force in the field of education \cite{Abdelghani2023,Ji2023,MacNeil2022}. The integration of MLLMs in science educational settings holds the promise of personalized and adaptive learning experiences, matching the needs of learners by delivering content that is both accessible and engaging. 

MLLMs are characterized by their ability to analyze and interpret data from different sources, such as spoken language, written text, or images. A key advantage of MLLMs over static sources of information, such as textbook representations, is the adaptive nature of these models. Both educators and students can use MLLMs to independently transform information into other modalities, such as textual information on the evolution of ant populations, into visual representations. This capability of MLLMs can be called "adaptive transformation." It allows knowledge representations to alter form flexibly, especially between text and images, but also within a modality, for example, to simplify complex images or make difficult text more understandable. This shows that MLLMs can change and adapt knowledge representations with great efficiency. One central question arising is how MLLMs can be used to foster knowledge acquisition and scientific competencies. To guide this question, we proposed a framework that is rooted in the Theory of Multimedia Learning and integrates MLLMs into multimodal learning (cf. section 3): Given visual or textual input and the user's requirements for adaptiveness and personalization, the MLLM produces textual and/or visual outputs. The MLLM does this by transforming content from text to image or vice versa or shifting from uni- to multimodality by adding a modality. Depending on the setting, the user of the MLLM can be either the educator or the learner itself. The framework introduced in this paper proved beneficial in describing and analyzing various scenarios of applying MLLM in (science) education, including content creation, fostering student learning in diverse contexts, and providing assessment and feedback. 


Whereas in the past, support had to be provided by the learning material or the educator, it is now possible for learners to generate such support adaptively. For example, with MLLMs it is possible for learners to summarize texts, highlight important text passages, or display similarities between related texts and images as key points. In addition, an adaptive use of MLLMs would allow learners to integrate selection aids into multimodal representations only when they need them. This would make it possible to avoid expertise reversal effects~\cite{Kalyuga2007}.

A problem with employing MLLMs for (self-directed) learning might be that learners require a high degree of self-regulatory skills (e.g., \citet{bannert2014process}) in order to effectively and constructively use MLLMs for learning purposes.
As mentioned earlier, learners can effectively use MLLMs to integrate multimodal learning materials, as MLLMs allow learners to select, organize, and integrate learning materials according to their needs from multimodal sources. This improves the overall access to as well as the preparation of the learning material. In addition, multimodal representations have the potential to help learners better grasp the concepts to be learned. For example, if learners are unable to understand complex texts or associate complex schematic representations, MLLMs can be used to adapt learning materials so that integration is possible. 
It is necessary to use MLLMs in a way that the generative activities are not completely absorbed by the AI, but proportionally also by the learners. Regulating this balance seems to be a central pedagogical challenge when using MLLMs in education.

Science learning is multimodal, requiring students to engage with a variety of information formats like diagrams, tabular data, natural-language texts, images, and animations and synthesize information as well as shift across various formats \cite{vanLeewen1990,doran2021teaching}. A lack of representational competency can hinder science learning \cite{Nitz2014}.
With the emergence of MLLMs, educators now have new potential tools to address the necessity for multimodality in science education. MLLMs are equipped to process and generate multimodal content for different levels of competency. This capability is crucial in a discipline where understanding often hinges on the ability to interpret complex diagrams, analyze diverse datasets, and synthesize information from multiple sources. With MLLMs, there is potential for enhanced guidance and support in both mastering individual modalities and transitioning fluidly between them. 

In summary, MLLMs can support learners in exploiting the potential of multimodal learning materials and presentations. In particular, through adaptive design options, either by the learners themselves or by the educators in advance, it is possible to support learners in the selection, organization and integration of information contained in text, images or other modalities. This adaptivity makes learning processes potentially more effective. For example, learners with a lot of prior knowledge on a specific topic can choose more complex representations, while learners with fewer skills can adapt learning materials to their needs. However, this also highlights the difficulty of regulating the effective use of MLLMs, either by the learners themselves or by the scripted use of AI-based applications to enable generative learning processes on the part of the learners. These issues provide an area for future research and numerous potential applications.

\section{Future Implications}
MLLMs could take the advantages of multimodal learning environments to the next level. Unlike conventional settings, such as hypertext environments, MLLMs offer a superior degree of interactivity, enabling thus a  more dynamic and engaging learning experience. 

One of the most notable advantages of MLLMs is their ability to swiftly transition between different modes upon request, a feature not typically feasible in traditional multimedia learning environments. This capability paves the way for highly individualized learning experiences, tailored to the unique needs and preferences of each learner. By combining the strengths of LLMs — such as interactivity and real-time generation of content — with the efficiency of multimodal learning in optimizing cognitive resource usage, MLLMs offer new ways to support learning.
Furthermore, MLLMs facilitate a seamless transition between learning modes, potentially enhancing thus the representational competencies of learners, which are crucial for a deep understanding of scientific concepts \cite{Nitz2014}. This seamless integration of different modalities can significantly improve the learning process.

The potential shift towards more responsive and personalized learning environments, as facilitated by MLLMs, represents a significant advancement in educational technology. This shift could lead to learning environments that are more attuned to the individual needs of students, thereby enhancing the overall learning experience. 

As MLLMs are still in their infancy, there is a need for further research, which should not only investigate the various potentials from personalized learning to inclusiveness and the ones mentioned in this article. It is also crucial to focus on the potential (further) shift of the educator’s role. As AI systems become multimodal, they potentially become more advanced, making it relevant to re-think and continuously re-evaluate the educator's role. It is crucial to understand that MLLMs, like any other AI system, should not replace the educator or, worse, be seen as competitors, but that MLLMs may alter the educators' role in the learning process \cite{Burbules2020,Schiff2020}. The educators’ role changes depending on the degree of automation. Different modes and degrees of implementation of AI systems are imaginable \cite{Molenaar2022}. However, the impact of MLLMs will not be limited to science education; for other disciplines, there seem to be analogous opportunities and challenges (cf. \citet{lee2023multimodality}). 

In light of these considerations, it becomes evident that the integration of MLLMs into educational frameworks across various disciplines necessitates a thoughtful and collaborative approach, where the evolving role of educators is recognized and supported, ensuring that these advanced technologies enhance, rather than overshadow, the human-centric aspect of teaching and learning.


\section{Declaration of AI and AI-assisted technologies in the writing process}
During the preparation of this work, the authors used ChatGPT (\href{www.chat.openai.com}{www.chat.openai.com}; GPT-3.5 and GPT-4) and Grammarly (\href{www.grammarly.com}{www.grammarly.com}) in order to improve the readability and language of single sentences. After using these tools, the authors reviewed and edited the content as needed and take full responsibility for the content of the publication.

\section{Declaration of competing interest}
The authors declare that they have no known competing financial interests or personal relationships that could have appeared to influence the work reported in this paper.

\bibliographystyle{apalike}
\bibliography{sample-base}
\end{document}